\documentclass[journal,onecolumn,12pt]{IEEEtran}

\usepackage{ifpdf}
\usepackage{cite}
\usepackage{graphicx}
\usepackage{amsmath}
\usepackage{algorithmic}
\usepackage{array}
\usepackage{lineno}
\usepackage{hyperref}
\usepackage{amssymb}
\usepackage{amsfonts}
\usepackage[linesnumbered,ruled,vlined]{algorithm2e}
\usepackage{subfigure}
\usepackage{color, colortbl}
\usepackage{float}
\usepackage{multicol}
\usepackage{url}
\definecolor{DarkGray}{gray}{0.9}
\definecolor{LightGray}{gray}{0.2}
\makeatletter
\g@addto@macro{\UrlBreaks}{\UrlOrds}
\makeatother
\begin{document}

\title{Low Cost Edge Sensing for High Quality Demosaicking}
%
%
%

\author{Yan~Niu et al.,
\thanks{The authors are  with  the  State  Key  Laboratory  of  Symbol  Computation and  Knowledge  Engineering,  Ministry  of  Education,  College  of  Computer Science and Technology, Jilin University, Changchun 130012, China (e-mail: niuyan@jlu.edu.cn;  ouyj@jlu.edu.cn). This work is supported by the National Natural Science Foundation of China under Grant NSFC-61472157 and NSFC-61170092.}}
\maketitle

\begin{abstract}
Digital cameras that use Color Filter Arrays (CFA) entail a demosaicking procedure to form full RGB images. As today's camera users generally require images to be viewed instantly, demosaicking algorithms for real applications must be fast. Moreover, the associated cost should be lower than the cost saved by using CFA. For this purpose, we revisit the classical Hamilton-Adams (HA) algorithm, which outperforms many sophisticated techniques in both speed and accuracy. Based on a close look at HA's strength and weakness, we design a very low cost edge sensing scheme. Briefly, it guides demosaicking by a logistic functional of the difference between directional variations. We extensively compare our algorithm with 28 demosaicking algorithms by running their open source codes on benchmark datasets. Compared to methods of similar computational cost, our method achieves substantially higher accuracy; Whereas compared to methods of similar accuracy, our method has significantly lower cost. Moreover, on test images of currently popular resolution, the quality of our algorithm is comparable to top performers, whereas its speed is tens of times faster.

\end{abstract}


%
\IEEEpeerreviewmaketitle

\section{Introduction}

\begin{figure}[h]
	\centering
		\includegraphics[width=0.25\linewidth]{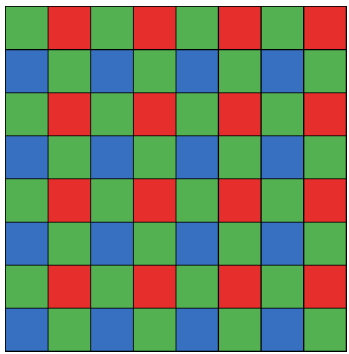}
	\caption{The most widely used Bayer pattern for CFA arrangement.}
	\label{fig:BayerCFA}
\end{figure}

The vast majority of current consumer digital cameras have Color Filter Arrays (CFA) placed on the light sensing units, to capture only one of the three primary color components at each pixel \cite{szeliski2010}. Fig.\ref{fig:BayerCFA} shows the most frequently used CFA layout named Bayer Pattern: in each $2\times 2$ subblock, the diagonal sensing units response to the green wavelength component and the anti-diagonal ones response to the red and blue wavelength components of light rays. Recovering the missing primary color values to form standard RGB color images, is called Demosaicking. 

A faithful demosaicking algorithm not only enables obtaining good quality images with low hardware cost, but also provides a potential solution to image compression. Therefore, demosaicking has been of intense interest in both academic research and industry. Similar to many ill-posed image recovery problems, what demosaicking research really aims to solve is demosaicking at non-regular regions such as edge and textures. Thus the numerous existing demosaicking methods commonly focus on how to accurately detect the least variation direction from the CFA sampled image data. They differ in the aspects of: 1) the domain to conduct finite differencing; 2) the measurement of directional variation by differencing; 3) the strategy of steering interpolation along local dominant direction; 4) the exploitation of the inter- and intra- channel correlation for higher accuracy; 5) the procedure to enhance the quality of a fully interpolated RGB image.
    
\textbf{Differencing Domain}\quad As a natural indicator of variation, magnitude of the first- or second-order finite differencing can be computed in each CFA sampled channel (e.g., \cite{HA96}\cite{Universal16}\cite{CCCA14} by Shao-Rehman); or across different channels (e.g. \cite{MSG13}\cite{PID16}). Alternatively, differencing can be conducted in each tentatively estimated color channel (e.g., \cite{HQLI04}), or their color-difference planes (e.g., \cite{LDINAT11}\cite{GBTF2010}). Many recent works perform differencing in the \emph{residual planes}, which are the difference between the CFA samples and intermediately interpolated channels, and have shown promising results (e.g., \cite{RI13}\cite{ARI15}\cite{ICC14}\cite{DDR16}). 

\textbf{Measuring Variation}\quad To measure directional variation, the Hamilton-Adams (HA) method combines the first- and second-order differencing magnitude in two channels at the current single pixel \cite{HA96}. This method is also adopted in subsequent works such as \cite{PID16} and \cite{VCD06}. A more robust approach is to accumulate the directional differencing magnitude in a local neighbourhood (e.g., \cite{SSD09}\cite{RI13}\cite{MSG13}\cite{DDR16}), or further at a mixture of scales (e.g., \cite{ASMS14}). 

\textbf{Edge Directed Interpolation}\quad HA compares the horizontal and vertical variation and selects the smoother direction to perform interpolation. In case of ties, interpolations in both directions are averaged \cite{HA96}. Su extended this idea by fusing the horizontal and vertical interpolation using machine learned weights \cite{WECD06}; Chung-Chan selected the local dominant direction based on the variance of directional color differences \cite{VCD06};  In \cite{Vote14}, local dominant direction is determined by voting for the horizontal or vertical or none edge hypotheses; Wu et al. relaxed the strict prerequisite for none-edge judgment to approximate equality \cite{PID16}. More methods estimate missing values by weighted summation of the estimation from the north, south, west and east directions, where the weights are obtained from the directional variation (e.g.\cite{SSD09}\cite{RI13}\cite{MSG13}\cite{DDR16}\cite{FDRI16}) and spatial distance (e.g.\cite{Universal16} \cite{BFDD17}). Ref.\cite{PCSD04} tests multiple direction hypotheses and chooses the one that shows the highest consistency among all channels. Deciding edge direction by maximizing a posteriori is also adopted in \cite{DFAPD07} and \cite{SSSC14}. 

\textbf{Exploitation of Cross-Channel Correlation}\quad The color channels of a natural image generally show strong correlation, meaning that the color or edge information of one channel is also implied by other channels. Hence cross-channel priors are extensively explored for demosaicking. For example, HA assumes that the color difference planes are locally bilinear. As this assumption fails at edges, Ref.\cite{ICC14} proposes compensating inter-channel interpolation by intra-channel interpolation if evidence of non-linearity presents; Ref.\cite{PCSD04} assumes that the three channels have consistent edge directions; Regularization is investigated to formulate the inter- and intra- channel correlation (e.g., \cite{RA08} \cite{LDINAT11}). Ref.\cite{MLRI16} assumes that the local region in one channel is a linear transform of another channel in the intermediate estimation step. 

\textbf{Quality Enhancement}\quad Due to the existence of cross-channel correlation, many works refine each channel by other channels' reconstruction alternatively and iteratively (e.g., \cite{AP02}\cite{SP05}\cite{AHD05}\cite{WECD06}\cite{OSAP10}\cite{IRI15}). Postprocessing techniques such as non-local regularization and median filtering have also been widely employed to suppress spurious high frequency components \cite{NMPM03}\cite{SSD09}\cite{LDINAT11}\cite{SSSC14}. It is known that non-local regularization and median filtering are essentially a series of iterative linear filtering, robust to outliers but computationally heavy \cite{niu12}. For efficiency, Wu et al. proposed a postprocessing technique based on machine learned regression priors.

\textbf{Deep Learning Demosaicing}\quad Convolutional Neural Networks (CNN) recently have attracted the attention of demosaicing research (\cite{MNN14}\cite{DJ16}\cite{DRL17}\cite{MDFCN18}). Some of these works have achieved the top demosaicing accuracy on benchmarks yet is faster than many classical methods (e.g., \cite{DJ16}). An implicit cost for CNN is the non-trivial memory required to store the trained model (e.g., \cite{DJ16},\cite{MDFCN18}) .

The accuracy of recent demosaicking methods keeps increasing, and so is the associated computational cost (see Sec.\ref{sec:experiment}). For real time visualization, sophisticated demosaicking may entail expensive processing hardware and high power supply, against the intention of using CFAs. In the vast literature on demosaicing, the HA method is extremely simple. Seemingly such simplicity may only yield baseline accuracy. However, it performs surprisingly well. Buades et al. tested the HA algorithm and $8$ well-known methods on $10$ images selected from McM dataset \cite{PCSD04}, and the HA algorithm is shown to achieve the least Mean Square Error (MSE) \cite{SSD09}. Gharbi et al. compared the HA algorithm with $16$ high-impact demosaicking methods of the literature (till year of 2016) \cite{DJ16} . While HA is the second fastest (only slower than Bilinear Interpolation), its Peak Signal to Noise Ratio (PSNR) accuracy on benchmark datasets Kodak \cite{Kodak} and McM is higher than $10$ methods. 

Although the HA algorithm does not put much effort on edge detection, it is highly effective. Based on a close look at the HA algorithm, we propose a high-quality fast edge-sensing image demosaicking scheme that adopts the HA pipeline. Particularly, we recover the green channel first, and then the green-red and green-blue color difference planes. For adaptive edge-sensing, we replace HA's green channel selective directional interpolation by blending the directional estimation, using a logistic functional of the difference between directional variations. We extend this edge-sensing strategy to the green-red and green-blue colour difference planes. This extension is not straightforward, since Bayer CFA samples the green channel twice as many the red or blue channels. Our approach is to derive a logistic functional to blend the diagonal and anti-diagonal estimation, leveraging the diagonal symmetry of the Bayer pattern. Then the green channel interpolation scheme is applicable to computing the rest missing values in the green-red and green-blue difference planes. The proposed demosaicking process is highly parallelable: although the red and blue channels have to be estimated subsequently to the green channel estimation, the restoration in each step at a pixel is independent of the restoration of other pixels. This feature means that our method is very suitable for Graphics Processing Units (GPU) and Field Programmable Gate Arrays (FPGA) implementation, achieving instant image visualization in real applications. 

The rest of the paper is organized as follows. Section-II analyzes the strength and weakness of the HA algorithm. Subsequently, Section \ref{sec:LHA} formulates a new fast edge-sensing demosaicking technique. Section \ref{sec:experiment} compares the efficiency and accuracy of our proposed method with the state-of-the-art methods by extensive experiments. Section \ref{sec:conclusion} concludes our work.

\section{Hamilton-Adams Demosaicking}  
\label{sec:HARevisit}

Assuming the mosaicked image $\mathbf{M}$ has $m$ rows and $n$ columns, let 
\[\mathcal{L}=\left\{(i,j)\in \mathbb{N}^2|i\in [1 \quad m],j\in [1 \quad n]\right\}\] 
be the set of all pixel positions. According to the Bayer mosaicking pattern (Fig.\ref{fig:BayerCFA}), we define
\begin{eqnarray*}
\mathcal{G}&=&\left\{(i,j)\in \mathbb{N}^2 |(i+j) \medspace \textrm{is even}\right\}\nonumber \\
\mathcal{R}&=&\left\{(i,j)\in \mathbb{N}^2 |i \medspace \textrm{is even}, j \medspace \textrm{is odd} \right\}\nonumber\\
\mathcal{B}&=&\left\{(i,j)\in \mathbb{N}^2 |i \medspace \textrm{is odd}, j \medspace \textrm{is even} \right\}
\end{eqnarray*}
to be the sets of positions where the green, red and blue values are originally available respectively. Hence their complementary sets $\mathcal{G}^{c}=\mathcal{L}\setminus \mathcal{G}$, $\mathcal{R}^{c}=\mathcal{L}\setminus \mathcal{R}$ and $\mathcal{B}^{c}=\mathcal{L}\setminus \mathcal{B}$ are the sets of positions where the green, red and blue values are to be recovered respectively. We use $r$,$g$ and $b$ to denote the original RGB components of a pixel $(i,j)$, and denote its estimated color components by adding a ``hat'' symbol to the corresponding notation.  For example, if a pixel $(i,j)\in \mathcal{G}$, we write its true RGB values as $\left(r,g,b\right)$ and its estimated RGB values as $\left(\hat{r},g,\hat{b}\right)$. 

\subsection{HA Green Channel Demosaicking}
Let $(i,j)\in \mathcal{G}^{c}$. The HA algorithm first computes its horizontal and vertical intensity variation, then selects the less variation direction to perform interpolation. In particular, at pixel $(i,j)$, the horizontal first and second order partial derivatives (denoted by $\partial_{h}$ and $\partial^{2}_{h}$), as well as the vertical first and second order differential (denoted by $\partial_{v}$ and $\partial^{2}_{v}$) of $\mathbf{M}$, are computed by 
\begin{eqnarray}
\partial_{h}\mathbf{M}(i,j) & = & \frac{\mathbf{M}\left(i,j+1\right)-\mathbf{M}\left(i,j-1\right)}{2} \nonumber \\
\partial^{2}_{h}\mathbf{M}(i,j)  & = & \frac{\mathbf{M}\left(i,j+2\right)+\mathbf{M}\left(i,j-2\right)-2\mathbf{M}\left(i,j\right)}{4}\nonumber\\
\partial_{v}\mathbf{M}(i,j)& = & \frac{\mathbf{M}\left(i+1,j\right)-\mathbf{M}\left(i-1,j\right)}{2} \nonumber \\
\partial^{2}_{v}\mathbf{M}(i,j) & = & \frac{\mathbf{M}\left(i+2,j\right)+\mathbf{M}\left(i-2,j\right)-2\mathbf{M}\left(i,j\right)}{4},
\label{eq:horiDiff}
\end{eqnarray}
Note that in Eq.\ref{eq:horiDiff}, pixels that are one unit away from $(i,j)$ are in $\mathcal{G}$, and pixels that are two units away are in the same set as $(i,j)$. 

The HA algorithm defines the horizontal variation $v_{h}$ and vertical variation $v_{v}$ as
\begin{eqnarray}
v_{h}& = & \left|\partial_{h}\mathbf{M}(i,j)\right|+\left|2\partial^{2}_{h}\mathbf{M}(i,j)\right|\nonumber\\
v_{v} & = & \left|\partial_{v}\mathbf{M}(i,j)\right|+\left|2\partial^{2}_{v}\mathbf{M}(i,j)\right|.
\label{eq:HA gradients}
\end{eqnarray}

Let $\bar{g}_{h}$ and $\bar{g}_{v}$ be the average of the neighbouring green values in the horizontal and vertical directions respectively, i.e.,
\begin{eqnarray}
\bar{g}_h & = & \frac{1}{2}\left(\mathbf{M}(i,j+1)+\mathbf{M}(i,j-1)\right)\nonumber\\
\bar{g}_v & = & \frac{1}{2}\left(\mathbf{M}(i+1,j)+\mathbf{M}(i-1,j)\right).
\label{eq:directional summation}
\end{eqnarray}

Finally, $\hat{g}(i,j)$ is estimated by
\begin{equation}
\hat{g}(i,j) = \left\{\begin{array}{ll}
               \bar{g}_h-\partial^{2}_{h}\mathbf{M}(i,j)& \textrm{if $v_{h} < v_{v} $}\\
							 \bar{g}_v-\partial^{2}_{v}\mathbf{M}(i,j)& \textrm{if $v_{h} > v_{v} $}\\
							\frac{1}{2}\left[\bar{g}_h-\partial^{2}_{h}\mathbf{M}(i,j)\right)+\left(\bar{g}_v-\partial^{2}_{v}\mathbf{M}(i,j)\right]
							& \textrm{if $v_{v} = v_{h}$}
                 \end{array}\right.
\label{eq:HA G}
\end{equation}
\subsection{HA Red and Blue Channel Demosaicking}
The HA demosaicking method utilizes the recovered green plane to regulate the blue and red recovery. Particularly for the Bayer CFA pattern, this is a typical $2\times 2$ times super-resolution problem, with available subsamples evenly spaced at every other row and column. Instead of directly enlarging the red plane $r(\mathcal{R})$ and blue plane $b(\mathcal{B})$, the HA algorithm enlarges the colour difference planes $\hat{g}(\mathcal{R})-r(\mathcal{R})$ and $\hat{g}(\mathcal{B})-b(\mathcal{B})$, based on the observation that $g(\mathcal{L})-r(\mathcal{L})$ and $g(\mathcal{L})-b(\mathcal{L})$ are generally smoother than $r(\mathcal{L})$ and $b(\mathcal{L})$ respectively. The magnification is simply performed by a bilinear interpolation 
\begin{subequations}
\begin{eqnarray}
\widehat{(g-r)}(i,j)= \sum_{(k,l)\in\mathcal{R}\cap \mathcal{N}_{i,j}} \omega_{k,l}\left(\hat{g}(k,l)-r(k,l)\right) \medspace \textrm {for $(i,j)\in \mathcal{R}^c$}
\label{eq: HA dgr}\\
\widehat{(g-b)}(i,j)= \sum_{(k,l)\in\mathcal{B}\cap \mathcal{N}_{i,j}} \varpi_{k,l}\left(\hat{g}(k,l)-b(k,l)\right) \medspace \textrm {for $(i,j)\in \mathcal{B}^c$},
\label{eq: HA dgb}
\end{eqnarray}
\end{subequations}
\normalsize
where $(k,l)$ index the pixels that are in the local neighbourhood $\mathcal{N}_{i,j}$ with $\hat{g}-r$ (in Eq.\ref{eq: HA dgr}) or $\hat{g}-b$ (in Eq.\ref{eq: HA dgb}) values available for bilinear interpolation; $\omega$ and $\varpi$ are the corresponding bilinear interpolation coefficients, determined by the spatial distance between $(i,j)$ and $(k,l)$.

Finally the missing red and blue values are recovered by
\begin{equation}
	\hat{r}(i,j)=\left\{\begin{array}{ll}
	g(i,j) - \widehat{(g-r)}(i,j) & \textrm {for $(i,j) \in \mathcal{G}$} \\
	\hat{g}(i,j) - \widehat{(g-r)}(i,j) & \textrm {for $(i,j) \in \mathcal{B}$}
	                    \end{array}
								\right. . 
\label{eq:HAR}
\end{equation}
and 
\begin{equation}
	\hat{b}(i,j)=\left\{\begin{array}{ll}
	g(i,j) - \widehat{(g-b)}(i,j) & \textrm {for $(i,j) \in \mathcal{G}$}\\
	\hat{g}(i,j) - \widehat{(g-b)}(i,j) & \textrm {for $(i,j) \in \mathcal{R}$}
	                    \end{array}
								\right. .
\label{eq:HAB}								
\end{equation}

\subsection {Advantages and limitations of the HA algorithm }
The high effectiveness of the HA method is due to the wisdom of taking full advantage of the green channel, which is sampled more densely than the other two channels. The green channel is recovered first based on available samples. It is then used to regulate the recovery of the red and blue channels. In other words, it trusts the sampling frequency more than edge detection. Such a strategy is suitable for today's digital CFA cameras, the resolution of which is generally several mega-pixels. This high sampling frequency means that the intensity at each pixel is highly correlated to its local neighbours; whereas to perform edge detection in a non-local neighbourhood, especially when $\frac{2}{3}$ information at each pixel is lost, could be time consuming. 

Nevertheless, HA's smoothness assumption is over simplified. In real applications, due to the existence of noise, the HA scheme restores the green component at a pixel exclusively from either its horizontal or vertical neighbours, as the variation in the two directions $v_{v}$ and ${v_{h}}$ are hardly equal (see Eq.\ref{eq:HA gradients} and Eq.\ref{eq:HA G}). This is disadvantageous in smooth regions, where more neighbours should be used to smooth out random noises. Moreover, its red and blue channel recovery assumes that the color difference plane is locally bilinear, which is seriously violated at edge or texture area and results in the ``false color'' artifacts \cite{NMPM03}. In the next section, we propose a more adaptive and flexible edge-sensing demosaicking scheme, which lifts the HA demosaicking accuracy to state-of-the-art comparable methods, while still being fast.

\section{Edge-Sensing Demosaicking by Logistic Functional of Difference between Directional Variations} 
\label{sec:LHA}

\subsection {Green Channel Demosaicking}
\label{subsec:greenReco}
The green channel demosaicking process of the HA algorithm, as shown in Eq.\ref{eq:HA G}, can be rewritten as
\begin{equation}
	\hat{g}(i,j) = \omega_{h}(\bar{g}_{h}-\partial^{2}_{h}\mathbf{M}(i,j))+(1-\omega_{h})(\bar{g}_{v}-\partial^{2}_{v}\mathbf{M}(i,j)),
	\label{eq:Unified}
\end{equation}
where 
\begin{equation}
	\omega_{h}=\left\{\begin{array}{ll}
	0 & \textrm{if $v_{h} > v_{v}$} \\
	1 & \textrm{if $v_{h} < v_{v}$}  \\
	\frac{1}{2} & \textrm{if $v_{h} = v_{v} $}.
	         \end{array}\right.
\label{eq:HAGWeight}								
\end{equation}  
   
In practice, even in very flat region, $v_{h}$ and $v_{v}$ are rarely equal because of noise. A more practical solution is to relax the strict equality requirement $v_{h}=v_{v}$ to the approximate equality $v_{h}\approx v_{v}$, which can be expressed by the inequality $\left|v_{h} - v_{v}\right|\leq T$, where $T$ is the allowed noise level. That is, 
\begin{equation}
	\omega_{h}=\left\{\begin{array}{ll}
	0 & \textrm{if $v_{h} - v_{v} > T $}  \\
	1 & \textrm{if $v_{h} - v_{v} < -T$}  \\
	\frac{1}{2} & \textrm{if $\left|v_{h} - v_{v}\right|\leq T$}.
	         \end{array}\right.
\label{eq:HAGWeightT}							
\end{equation}  
Although $\omega_{h}$ defined by Eq.\ref{eq:HAGWeightT} is more flexible than by Eq.\ref{eq:HAGWeight}, the value of $T$ has to be carefully defined for each image, as a small bias in $T$ may lead to an opposite interpolation decision. Desirably, $\omega_{h}$ should be a continuous function, which smoothly blends the estimation from both directions, thus a small bias does not cause the demosaicking result to vary abruptly. In particular, rather than using the step function defined by Eq.\ref{eq:HAGWeightT}, we seek for a smooth function $\omega_{h}$ that meets the criteria:
\begin{enumerate}
	\item $\omega_{h}\to 0$, when $T < v_{h} - v_{v} \to \infty$;
	\item $\omega_{h}\to 1$, when $-T > v_{h} - v_{v} \to -\infty$;
	\item $\omega_{h}\approx \frac{1}{2}$, if $\left|v_{h} - v_{v}\right| \leq T$;
\end{enumerate}
Note that, $1-\omega_{h}$ and $\omega_{h}$ should have the same form. That is, if there is a function $f$, such that $\omega_{h}=f(v_{h} - v_{v})$, then $1-\omega_{h}=f(v_{v} - v_{h})$ should hold. In other words, 
	\begin{equation}
f(v_{h} - v_{v}) + f(v_{v} - v_{h}) = 1.		
	\end{equation}
It can be shown that the logistic function
\begin{equation}
	f_{k}(x) = \frac{1}{1+e^{kx}},
\end{equation}
where $k$ is a positive real number adjusting the convergence of $f_{k}(x)$, fulfills all requirements on $\omega_{h}$. Thus we define
\begin{equation}
	\omega_h  =  \frac{1}{1+e^{k\left(v_{h} - v_{v}\right)}} .
	\label{eq:LogisticFunction}
\end{equation} 
It can be verified that
\begin{equation}
   1-\omega_h = \frac{1}{1+e^{k\left(v_{v} - v_{h}\right)}}.
\end{equation} 

\begin{figure}[ht!]
\centering
	\includegraphics[width=0.75\textwidth,height=0.25\textheight]{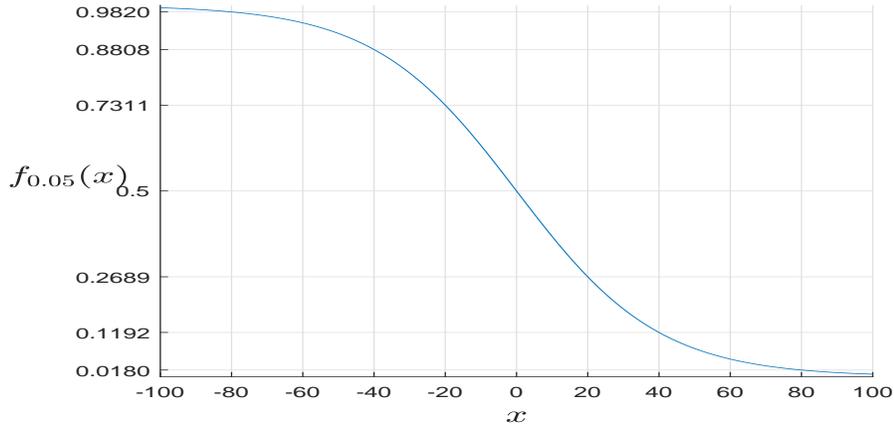}
	\caption{The curve of logistic function $f_{k}(x)=(1+e^{kx})^{-1}$, with $k=0.05$, which we use to balance the contribution from the horizontal and vertical neighbours.}
	\label{fig:logisticFuncCurv}
\end{figure}

Algorithm \ref{alg:green interpolation} summarizes the green channel demosaicking pipeline. To examine the influence of hyper-parameter $k$ on demosaicking performance, we run the algorithm on $100$  high quality natural images from Waterloo Exploration Database \cite{Waterloo17}\cite{Waterloo}, with $k$ varying from $0.01$ to $1.0$ at a step size of $0.01$. We observe that $k=0.05$ yields the highest PSNR (averaged over the $100$ training images), hence we fix $k$ to be $0.05$ in this work. Fig.\ref{fig:logisticFuncCurv} plots the function curve of $f_{0.05}(x)$. Note that, the high pass filtering involved in the interpolation scheme does not preserve energy. Consequently, $\hat{g}$ might be out of the range of $[\min(g(\mathcal{G})),\max(g(\mathcal{G}))]$, hence we clip such $\hat{g}$ values to be either $\min(g(\mathcal{G}))$ or $\max(g(\mathcal{G}))$, whichever is closer, at the final step of the algorithm.

\begin{algorithm}
\KwIn{Mosaicked image $\mathbf{M}$; hyper-parameter $k$}
\KwOut{ $\hat{g}(i,j)$ for each $(i,j) \in \mathcal{G}^c$}
$m_{max} := \max(g(\mathcal{G}))$;\\
$m_{min} := \min(g(\mathcal{G}))$;\\
\For{each $(i,j)\in \mathcal{G}^c$}
{
  Compute $\partial_{h}$,$\partial_{v}$,$\partial^2_{h}$,$\partial^2_{v}$ of $\mathbf{M}$ at pixel $(i,j)$ by Eq.\ref{eq:horiDiff};\\
	Compute $v_{h}$ and $v_{v}$ by Eq.\ref{eq:HA gradients}, $\bar{g}_{h}$ and $\bar{g}_{v}$ by Eq.\ref{eq:directional summation};\\
	Compute $\omega_{h}$ by Eq.\ref{eq:LogisticFunction};\\
	Compute $\hat{g}(i,j)$ by Eq.\ref{eq:Unified};\\
	$\hat{g}(i,j) = \max(\min(\hat{g}(i,j),m_{max}),m_{min})$;
}

\Return{$\hat{g}(\mathcal{G}^c)$}
\caption{Green Channel Demosaicking}
\label{alg:green interpolation}
\end{algorithm}

\subsection {Red and Blue Channels Demosaicking}

We transform $r(\mathcal{R}^c)$ and $b(\mathcal{B}^c)$ estimation to $(\hat{g}-r)(\mathcal{R}^c)$ and $(\hat{g}-b)(\mathcal{B}^c)$ interpolation. We treat the two channels in the same fashion, hence this section only articulates the red channel demosaicking. Its blue channel counterpart can be derived by simply exchanging the positions of red and blue in the algorithm. 

To respect edges and textures, we apply our edge-sensing strategy also to the red channel. This is cannot be done by a straightforward extension from the green to the red channel. Due to Bayer CFA color sensors arrangement, in the green channel, at a pixel $(i,j) \in \mathcal{G}^c$, its horizontal and vertical neighbours all have original green values available. In contrast, if $(i,j) \in \mathcal{R}^c$, at most two of its horizontal and vertical neighbours have green-red difference values available. Our approach is to leverage the diagonal symmetry of the red sensors' positions. We first derive the edge-sensing interpolation scheme for $(g-r)(i,j)$, where $(i,j)\in \mathcal{B}$, using its diagonal and anti-diagonal neighbours. This makes the green-red difference values available at the horizontal and vertical neighbours for each of the rest pixels. We then infer $r(\mathcal{G})$ from $(\hat{g}-r)(\mathcal{R})$ and the estimated $(g-r)(\mathcal{B})$. 

\subsubsection {Estimating red values at $\mathcal{B}$} 
\begin{figure*}[h!]
    \centering
    \subfigure[]{
				\label{fig:grAtB}
        \includegraphics[width=0.45\linewidth]{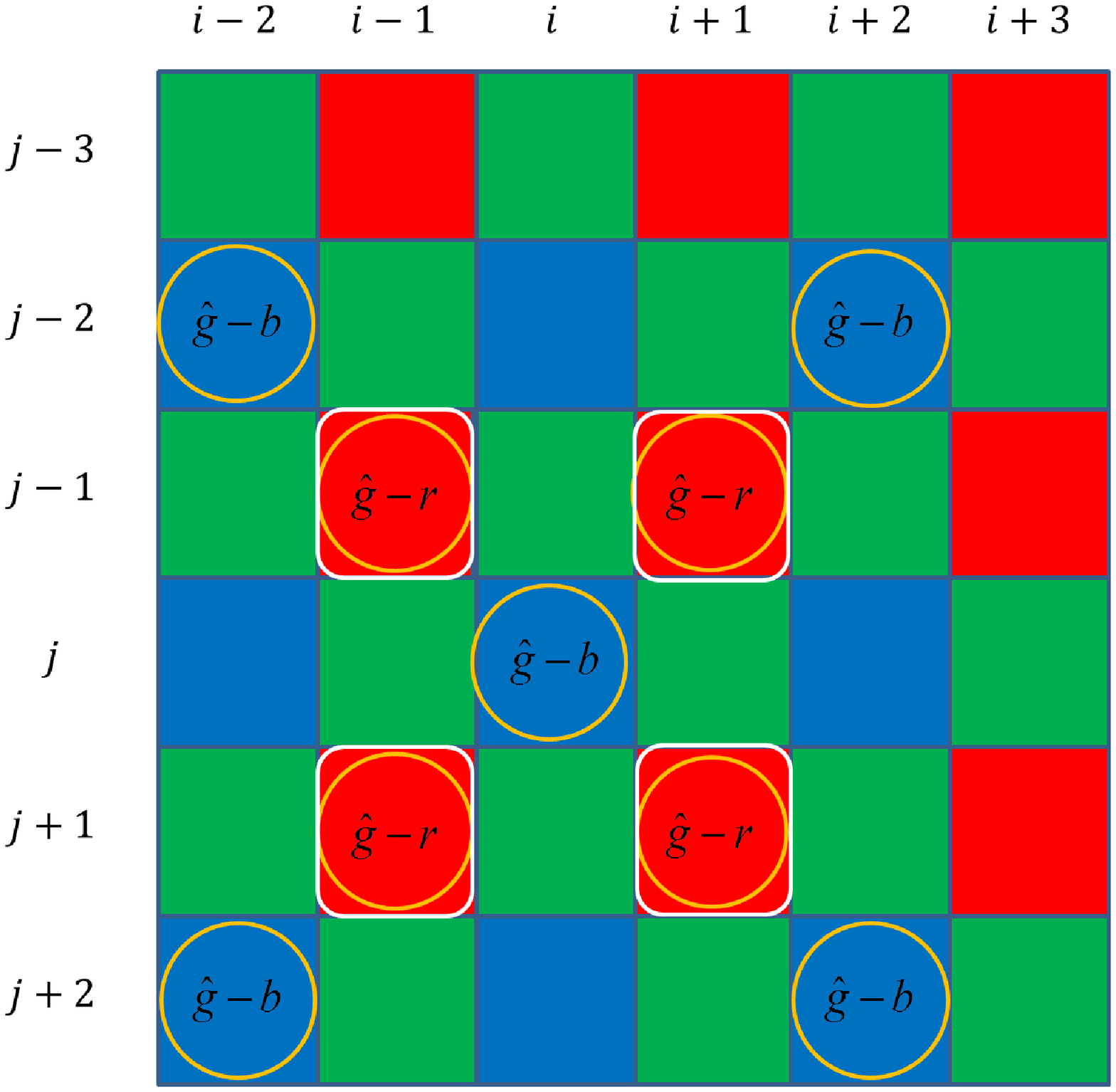}
    }
    ~ 
    \subfigure[]{
				\label{fig:grAtG}
        \includegraphics[width=0.45\linewidth]{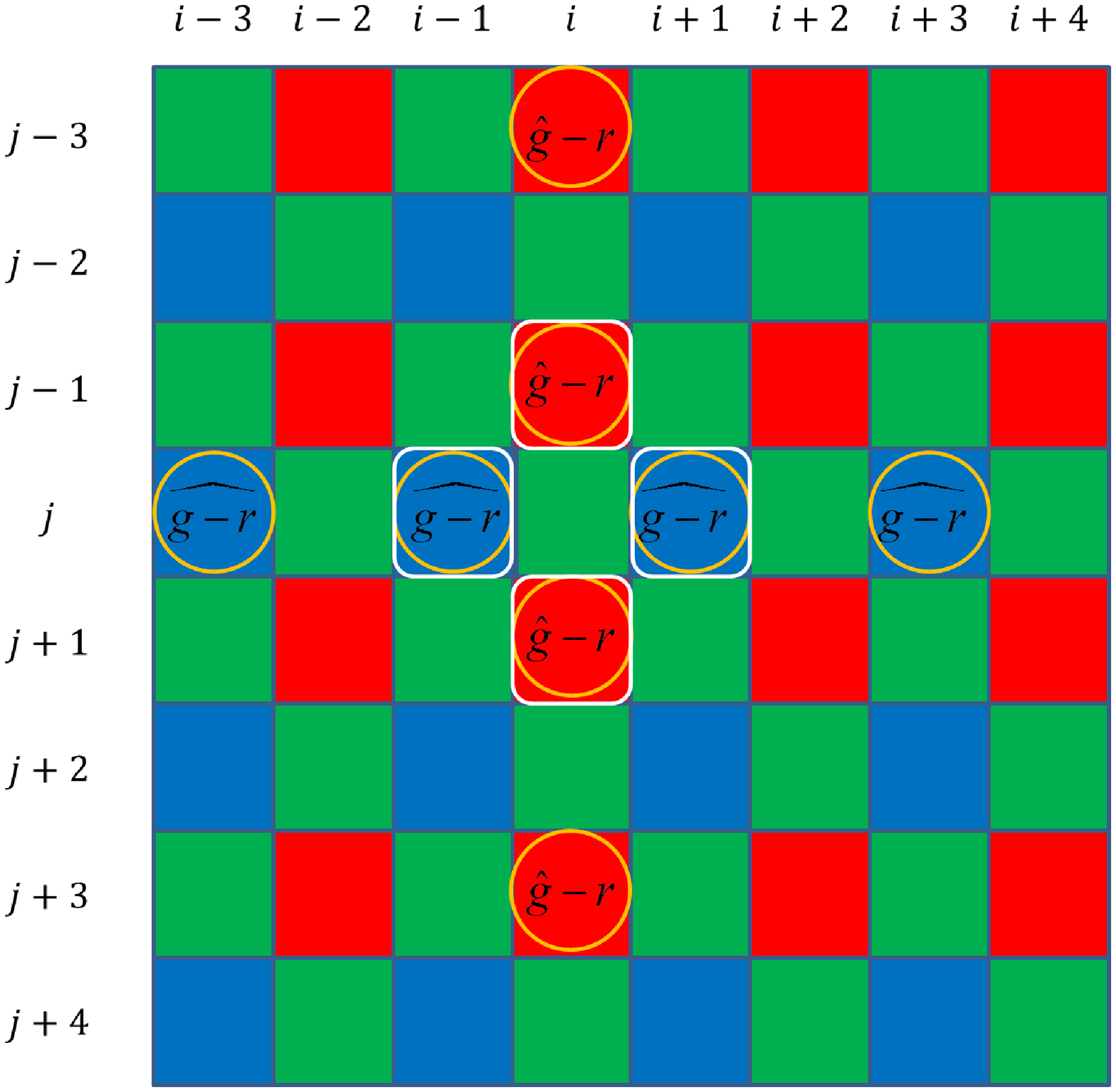}
		}
    \caption{Illustration of recovering $(g-r)(i,j)$ for $(i,j)\in \mathcal{R}^c$. Pixels used for computing the second order partial derivatives are surrounded by a circle; Pixels used for computing the first order partial derivatives are surrounded by both a circle and a curved square. (a) First, for each pixel $(i,j) \in\mathcal{B}$, $\widehat{(g-r)}(i,j)$ is obtained from its diagonally and anti-diagonally neighbouring $(\hat{g}-r)$ and $(\hat{g}-b)$ values; (b) Subsequently, at each pixel $(i,j) \in\mathcal{G}$, $\widehat{(g-r)}(i,j)$ is obtained from its vertically neighbouring $(\hat{g}-r)$ values and horizontally neighbouring $\widehat{(g-r)}$ values.}
\end{figure*}

As shown in Fig.\ref{fig:grAtB}, the nearest available red values around a pixel $(i,j) \in \mathcal{B}$ are $r(i-1,j-1)$, $r(i+1,j+1)$, $r(i-1,j+1)$, and $r(i+1,j-1)$, located in the diagonal and anti-diagonal directions. To obtain edge information, we compute the difference between the diagonal and anti-diagonal intensity variation (in the mosaicked image plane $\mathbf{M}$). We then use the logistic function value of this difference to weight the contribution of $(\hat{g}-r)$ at $(i-1,j-1)$, $(i+1,j+1)$, $(i-1,j+1)$, and $(i+1,j-1)$ to restore $(\hat{g}-r)(i,j)$. 

In particular, we compute the first and second order diagonal and anti-diagonal partial derivatives of $\mathbf{M}$ at $(i,j)$ by 
\begin{eqnarray}
\partial_{d}\mathbf{M}(i,j) & = & \frac{\mathbf{M}\left(i+1,j+1\right)-\mathbf{M}\left(i-1,j-1\right)}{2\sqrt{2}} \nonumber \\
\partial^{2}_{d}\mathbf{M}(i,j) & = & \frac{\mathbf{M}\left(i+2,j+2\right)+\mathbf{M}\left(i-2,j-2\right)-2\mathbf{M}\left(i,j\right)}{8}\nonumber\\
\partial_{a}\mathbf{M}(i,j) & = & \frac{\mathbf{M}\left(i-1,j+1\right)-\mathbf{M}\left(i+1,j-1\right)}{2\sqrt{2}} \nonumber \\
\partial^{2}_{a}\mathbf{M}(i,j) & = & \frac{\mathbf{M}\left(i-2,j+2\right)+\mathbf{M}\left(i+2,j-2\right)-2\mathbf{M}\left(i,j\right)}{8},
\label{eq:diagDiff}
\end{eqnarray}
\normalsize
The local intensity variation in the diagonal and anti-diagonal directions are computed as
\begin{eqnarray}
v_{d}& = & \left|\partial_{d}\mathbf{M}(i,j)\right|+\left|2\sqrt{2}\partial^{2}_{d}\mathbf{M}(i,j)\right|\nonumber\\
v_{a} & = & \left|\partial_{a}\mathbf{M}(i,j)\right|+\left|2\sqrt{2}\partial^{2}_{a}\mathbf{M}(i,j)\right|.
\label{eq:diagVari}
\end{eqnarray}
Let $\omega_{d}$ be the logistic function value of $v_{d}-v_{a}$, i.e.,
\begin{equation}
	\omega_d  =  \frac{1}{1+e^{k\left(v_{d} - v_{a}\right)}} ,
	\label{eq:diagWeight}
\end{equation} 
where hyper-parameter $k$ is fixed to $0.05$, as described in Section \ref{subsec:greenReco} for green channel recovery.

Define 
\scriptsize
\begin{eqnarray}
\overline{(g-r)}_{d} & = & \frac{(\hat{g}-r)\left(i+1,j+1\right)+(\hat{g}-r)\left(i-1,j-1\right)}{2} \nonumber \\
\overline{(g-r)}_{a} & = & \frac{(\hat{g}-r)\left(i-1,j+1\right)+(\hat{g}-r)\left(i+1,j-1\right)}{2},
\label{eq:diagMean}
\end{eqnarray}
\normalsize
which compute the diagonal mean and anti-diagonal mean of $(\hat{g}-r)$ at $(i,j)$. 
Furthermore, the second order partial derivatives in the colour-difference plane $\hat{g}-b$ at $(i,j)$ are approximated by the central differencing scheme, 
\begin{eqnarray}
\partial^2_{d}(\hat{g}-b) & = & \frac{(\hat{g}-b)\left(i+2,j+2\right)+(\hat{g}-b)\left(i-2,j-2\right)-2(\hat{g}-b)\left(i,j\right)}{8} \nonumber \\
\partial^2_{a}(\hat{g}-b) & = & \frac{(\hat{g}-b)\left(i+2,j-2\right)+(\hat{g}-b)\left(i-2,j+2\right)-2(\hat{g}-b)\left(i,j\right)}{8},
\label{eq:diagMean}
\end{eqnarray}
\normalsize

We infer $(g-r)(i,j)$ by fusing the directional estimation using $\omega_{d}$, that is, 
\begin{equation}
	\widehat{(g-r)}(i,j) = \omega_{d}\left(\overline{(g-r)}_{d}-\partial^2_{d}(\hat{g}-b)\right)+(1-\omega_{d})\left(\overline{(g-r)}_{a}-\partial^2_{a}(\hat{g}-b)\right),
	\label{eq:grAtB}
\end{equation}
\normalsize
which recovers $r(i,j)$ by  
\begin{equation}
	\hat{r}(i,j) = \hat{g}(i,j)-\widehat{(g-r)}(i,j),
	\label{eq:rAtB}
\end{equation}
for $(i,j)\in\mathcal{B}$.

\subsubsection {Estimating red values at $\mathcal{G}$} 
Once $\widehat{(g-r)}(\mathcal{B})$ is available, $\widehat{(g-r)}(\mathcal{G})$ can be estimated from its horizontal and vertical neighbours, as shown in Fig. \ref{fig:grAtG}. Note that in this step, for each $(i,j) \in \mathcal{G}$, either $(\hat{g}-r)$ or $\widehat{(g-r)}$ values have been already computed at the four nearest neighbours $(i-1,j)$,$(i+1,j)$,$(i,j-1)$ and $(i,j+1)$. For notation simplicity, we denote them uniformly by $\widetilde{(g-r)}$. In the green-red difference plane at pixel $(i,j)$, we compute its horizontal and vertical average values $\overline{(g-r)}_{h}$ and $\overline{(g-r)}_{v}$ by 
\begin{eqnarray}
\overline{(g-r)}_{h} & = & \frac{\widetilde{(g-r)}\left(i,j+1\right)+\widetilde{(g-r)}\left(i,j-1\right)}{2} \nonumber \\
\overline{(g-r)}_{v} & = & \frac{\widetilde{(g-r)}\left(i+1,j\right)+\widetilde{(g-r)}\left(i-1,j\right)}{2}.
\label{eq:colorDiffHoriMean}
\end{eqnarray}
\normalsize
Moreover, we approximate the second order partial derivatives of $\widetilde{(g-r)}$ at $(i,j)$ by the central differencing scheme as 
\scriptsize
\begin{eqnarray}
\partial^2_{h}\widetilde{(g-r)} & = & \frac{\partial_{h}\widetilde{(g-r)}\left(i,j+1\right)-\partial_{h}\widetilde{(g-r)}\left(i,j-1\right)}{2} \nonumber \\
\partial^2_{v}\widetilde{(g-r)} & = & \frac{\partial_{v}\widetilde{(g-r)}\left(i+1,j\right)-\partial_{v}\widetilde{(g-r)}\left(i-1,j\right)}{2},
\label{eq:colorDiffCurv}
\end{eqnarray}
\normalsize
where $\partial_{h}\widetilde{(g-r)}(i,j-1)$, $\partial_{h}\widetilde{(g-r)}(i,j+1)$, $\partial_{v}\widetilde{(g-r)}(i-1,j)$ and $\partial_{v}\widetilde{(g-r)}(i+1,j)$
are further approximated by central differencing
\begin{eqnarray}
\partial_{h}\widetilde{(g-r)}(i,j-1) & = & \frac{\partial_{h}\widetilde{(g-r)}(i,j+1)-\partial_{h}\widetilde{(g-r)}(i,j-3)}{4} \nonumber \\
\partial_{h}\widetilde{(g-r)}(i,j+1) & = & \frac{\partial_{h}\widetilde{(g-r)}(i,j+3)-\partial_{h}\widetilde{(g-r)}(i,j-1)}{4} \nonumber \\
\partial_{v}\widetilde{(g-r)}(i-1,j) & = & \frac{\partial_{h}\widetilde{(g-r)}(i+1,j)-\partial_{h}\widetilde{(g-r)}(i-3,j)}{4} \nonumber \\
\partial_{v}\widetilde{(g-r)}(i+1,j) & = & \frac{\partial_{h}\widetilde{(g-r)}(i+3,j)-\partial_{h}\widetilde{(g-r)}(i-1,j)}{4} \nonumber \\
\label{eq:colorDiffGrad}
\end{eqnarray}

\normalsize
Subsequently, $\widehat{(g-r)}(i,j)$ is given by 
\begin{equation}
	\widehat{(g-r)}(i,j) = \omega_{h}\left(\overline{(g-r)}_{h}-\partial^2_{h}\widetilde{(g-r)}\right)+(1-\omega_{h})\left(\overline{(g-r)}_{v}-\partial^2_{v}\widetilde{(g-r)}\right),
	\label{eq:grAtG}
\end{equation}
\normalsize
where $\omega_{h}$ is computed by the same formula as in Eq.\ref{eq:horiDiff}, Eq.\ref{eq:HA gradients} and Eq.\ref{eq:LogisticFunction}.
Finally, $r(i,j)$ is restored by Eq.\ref{eq:rAtB} for $(i,j)\in\mathcal{G}$. Algorithm \ref{alg:red interpolation} summarizes the estimation process of the missing red components.
\begin{algorithm}
\KwIn{Mosaicked image $\mathbf{M}$; estimated green channel $\hat{g}(\mathcal{L})$; hyper-parameter $k$}
\KwOut{ $\hat{r}(i,j)$ for each $(i,j) \in \mathcal{R}^c$}
$r_{max} := \max(r(\mathcal{R}))$;\\
$r_{min} := \min(r(\mathcal{R}))$;\\
\For{each $(i,j)\in \mathcal{B}$}
{
   Compute $\hat{r}(i,j)$ by sequentially implementing Eq.\ref{eq:diagDiff} until Eq.\ref{eq:rAtB};\\
	 $\hat{r}(i,j) = \max(\min(\hat{r}(i,j),r_{max}),r_{min})$;
}
\For{each $(i,j)\in \mathcal{G}$}
{
  Compute $\widehat{(g-r)}(i,j)$ by sequentially implementing Eq.\ref{eq:colorDiffHoriMean} until Eq.\ref{eq:grAtG};\\
	Compute $\hat{r}(i,j)$ by Eq.\ref{eq:rAtB};\\
	$\hat{r}(i,j) = \max(\min(\hat{r}(i,j),r_{max}),r_{min})$;
}

\Return{$\hat{r}(\mathcal{R}^c)$}
\caption{Red Channel Demosaicking}
\label{alg:red interpolation}
\end{algorithm}

At image boundaries where pixels required for central differencing or averaging are unavailable, we simply restore the missing colour components by linear interpolation or nearest-neighbour interpolation.

\section{Experimental Results}
\label{sec:experiment}
We experimentally evaluate the proposed algorithm, which we name Logistic Edge-Sensing Demosaicking (LED). To examine the pure effectiveness of steering demosaicking by logistic functional of the difference between directional variation, we \emph{do not} enhance the image restoration quality by any post-processing or refinement technique. 

\textbf{Datasets} \quad Following the literature convention, we first test LED on traditional benchmarks Kodak \cite{Kodak} and McM \cite{LDINAT11}. The Kodak dataset contains 24 images of size $768\times 512$ and the McM dataset contains 18 images of size $500\times 500$. As each of these test images has fewer than $0.4$-Mega pixels, whereas current consumer camera resolution typically has several Mega pixels, experiments on traditional McM and Kodak datasets may not fully reflect real applications. To examine the potential performance of LED in real practice, we further test it on the $3072\times 2048$ (about $6.2$-Mega pixels) version of the Kodak dataset \cite{HRKodak}, the resolution of which is comparable to the $8$-Mega pixels resolution of Iphone6. We term this modern resolution Kodak dataset as MR Kodak.

\textbf{Comparison and Metrics} \quad Beside comparing to the HA algorithm, we extensively compare LED with 28 existing demosaicking methods by running their publicly available source codes. The performance comparison is conducted in terms of demosaicking accuracy and efficiency. We measure the accuracy of the demosaicked images by Peak Signal to Noise Ratio (PSNR), Structural SIMilarity (SSIM) and S-CIELAB \cite{SCIELAB}. In the case that the competing methods have source codes in MATLAB, we measure the demosaicking efficiency by timing the particular demosaicking process, which outputs the final RGB image from the Bayer mosaicked input, on McM $500\times 500$ images. More specifically, if the demosaicking process is implemented by a single function in the source code, we add the MATLAB timing function \textbf{timeit} to record its running time; Whereas if the demosaicking process consists of multiple functions, we use the MATLAB timing function \textbf{tic} and \textbf{toc}. In the case that the competing methods have source codes in C, we use the time library functions \textbf{clock}. 

Due to the ``warm up'' factor, the demosaicking generally takes longer on the first test image than the other images, hence we report the median running time from the 2nd to the 18th McM images as the final time measurement. All experiments are conducted on a 2.8GHz Intel i7-4900MQ CPU with 8GB RAM, unless otherwise specified.  

\textbf{Parameter Settings} \quad The only hyper-parameter of our method to set is the logistic function steepness coefficient $k$ in Eq.\ref{eq:LogisticFunction}, which is fixed to $0.05$ (see Section\ref{subsec:greenReco}) in all experiments. Many existing works shave off image boundaries of various width from measuring demosaicking accuracy (for example, $11$ pixels in Ref.\cite{LSLCD13} and $20$ pixels in Ref.\cite{LDINAT11}), and we also shave off $4$ pixels wide image boundaries. If the source code of a competing method does not specify the shave-off boundary width, we also set it to $4$. For methods that simultaneously address demosaicking and denoising, we set the additional noise level to zero in their source codes. We leave other parameters (including boundary shaved-off size) as their default values in the original code, since they may lead to the optimal performance. Nevertheless, we suggest that future research to take image boundaries into account, as the demosaicked image should not shrink in real applications.   

\subsection{Numerical Evaluation on Low Resolution Images}
Table \ref{tab:IndividualKodakLR} and Table \ref{tab:IndividualMcM} present the demosaicking accuracy, quantitatively measured by PSNR for each channel, PSNR for the whole image (a.k.a., cPSNR), SSIM and S-CIELAB, of the proposed LED algorithm on each individual image from Kodak and McM respectively. Table \ref{HAvsLED} compares the accuracy and computation time of the LED against the traditional HA method under the same implementation settings. Significantly, LED improves HA by $2.51$dB and $1.74$dB in PSNR, $0.01$ and $0.01$ in SSIM, $0.27$ and $0.31$ in S-CIELAB on Kodak and McM respectively, at an extra cost of merely $0.038$ seconds for $500 \times 500$ pixels. 
\begin{table*}[tp]
    \centering
				\label{tab:IndividualKodakLR}
        \includegraphics[height=0.9\textheight]{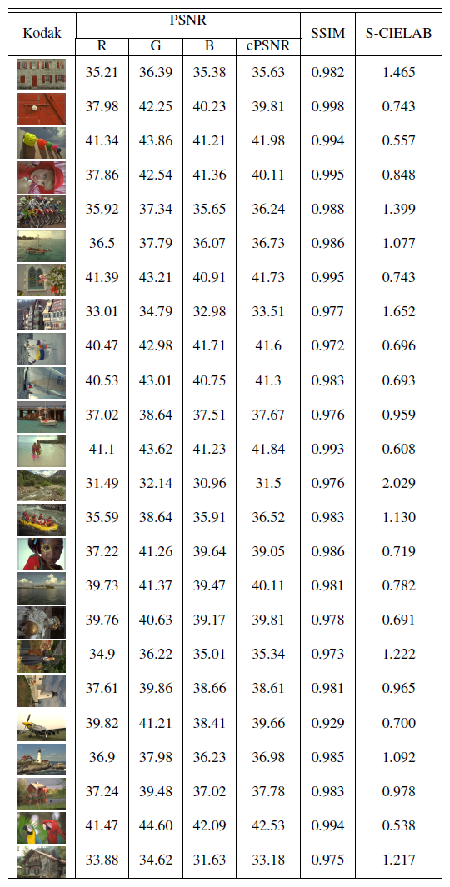}	
    \caption{Demosaicking accuracy of the proposed method LED, measured by PSNR of each channel, cPSNR, SSIM and S-CIELAB, on each individual image of the traditional low resolution Kodak dataset.}
\end{table*}

\begin{table*}[tp]
    \centering
				\label{tab:IndividualMcM}
        \includegraphics[height=0.9\textheight]{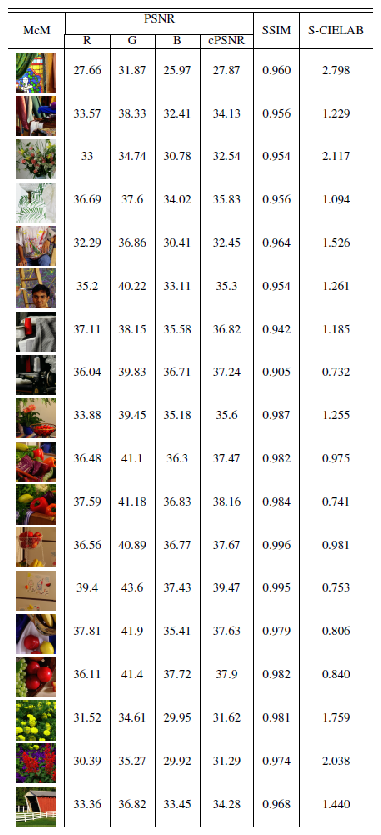}	
    \caption{Demosaicking accuracy of the proposed method LED, measured by PSNR of each channel, cPSNR, SSIM and S-CIELAB, on each individual image of the McM dataset.}
\end{table*}

\begin{table*}[tp]
\centering
\begin{tabular}{c|c|c|c|c|c|c|c}
\hline
\hline
 & \multicolumn{3}{|c|}{Kodak} & \multicolumn{4}{|c}{McM}\\
 \cline{2-8}
&  &  &  &  &  &  & time \\
& \raisebox{1.5ex}{cPSNR} & \raisebox{1.5ex}{SSIM} & \raisebox{1.5ex}{S-CIELAB} & \raisebox{1.5ex}{cPSNR} & \raisebox{1.5ex}{SSIM} & \raisebox{1.5ex}{S-CIELAB} &  {(sec)}\\
\hline
\cline{1-8}
  HA & $35.80$ &	0.971 & 1.246 & 33.49 & 0.958 & 1.622 & \textbf{0.024}\\
\hline
\cline{1-8}
 LED (ours) & \textbf{38.31} &	\textbf{0.98}2 &	\textbf{0.979} &	\textbf{35.23} & \textbf{0.968} & \textbf{1.308} & 0.062 \\
\hline
\end{tabular}
\vspace{1ex}
\caption{Performance comparison between the traditional HA and the proposed LED. Demosaicking accuracy is measured by cPSNR, SSIM and S-CIELAB values averaged over the Kodak and McM datasets; Demosaicking efficiency is measured by the median running time (in seconds) of demosaicking. Bold numbers indicate the superior performance under each metric.}
\label{HAvsLED}
\end{table*}

We compare LED with previous demosaicking methods by running their available source codes, mostly found according to \cite{codeList1} and \cite{CodeList2} \footnote{Deep Convolutional Neural Network based method proposed in \cite{DJ16} has Matlab code available online. However, as deep learning methods trade memories for computation time and accuracy, they are in a very different vein from our method, and hence Ref.\cite{DJ16} is not included in this experiment.}. They are: Alternating Projection (AP) \cite{AP02} (using the implementation by Y. M. Lu in \cite{OSAP10}); High Quality Linear Interpolation (HQLI) \cite{HQLI04} (using the MATLAB build-in function \textbf{demosaic}); Primary Consistent Soft Decision (PCSD) \cite{PCSD04}; Adaptive Homogeneity-Directed (AHD) \cite{AHD05};  Directional Linear Minimum Mean Square-Error Estimation (DLMMSE) \cite{DLMMSE05}; Weighted Edge and Color Difference (WECD) \cite{WECD06}; Total Least Square (TLS) \cite{TLS06}; Directional Filtering and A Posteriori Decision (DFAPD) \cite{DFAPD07}; Wavelet Analysis of Luminance Component (WALC) \cite{WALC07}; Heterogeneity-Projection Hard-Decision (HPHD) \cite{HPHD07}; Regularization Approach (RA) \cite{RA08}; Self-Similarity Driven (SSD) \cite{SSD09}; Contour Stencils (CS) \cite{CS12}; One Step Alternating Projections (OSAP) \cite{OSAP10}; Local Directional Interpolation and Non-local Adaptive Thresholding (LDINAT) \cite{LDINAT11}; Directional Filtering and Weighting (DFW) \cite{DFW12}; Residual Interpolation (RI) \cite{RI13}; Multiscale Gradient (MSG) \cite{MSG13}; Least Square Luma-Chroma Demultiplexing and Noise Estimation (LSLCD-NE)\cite{LSLCD13}; Flexible Image Processing Framework (FlexISP) \cite{FlexISP14} (using the implementation by Tan et al. in \cite{ADMM17}\footnote{The original implementation of FlexISP in \cite{ADMM17} computes PSNR for each channel first, then averages them as cPSNR. We modified this computation, by using the mean squared error over all pixels and all channels to compute cPSNR.}); Inter-color Correlation \cite{ICC14}; Adaptive Residual Interpolation (ARI) \cite{ARI15}; Multidirectional Weighted Interpolation (MDWI) \cite{MDWI15} (using the implementation found in Github); Directional Difference Regression (DDR) \cite{DDR16}; Minimized-Laplacian Residual Interpolation (MLRI) \cite{MLRI16}; Sequential Energy Minimization \cite{SEM16}; and Alternating Direction Method of Multipliers (ADMM) \cite{ADMM17}\footnote{Same modification as we did for FlexISP.}. Web addresses of these source codes are provided along with the bibliography of this work. 

For the clearance of comparison, Table \ref{tab:LEDvsOthers} only shows the accuracy measured by cPSNR and efficiency measured by running time of each competing method on the McM dataset, which is more challenging than the Kodak dataset \cite{FDRI16}. Evidently, it is observed that:
\begin{itemize}
	\item \emph{None} of the competing methods outperforms the proposed method by both higher cPSNR and lower computation cost; Whereas the proposed LED clearly outperforms $18$ out of $28$ methods by both cPSNR and running time. 
	\item OS-AP and LSLCD-NE have similar running time to the proposed method, but their cPSNRs are about $2$dB and $1.15$dB lower respectively. SSD and FlexIP have similar (slightly superior) cPSNRs to the proposed method, but they are about $69$ and $2633$ times slower.
	\item The proposed LED has lower cPSNR than RI, ICC, MLRI, CS, DDR, MDWI, ARI and LDINAT, but is about $12$, $17$, $27$, $28$, $122$, $280$, $420$ and $4400$ times faster than them respectively. 
\end{itemize}
\begin{table*}[tp]
  \centering
  
    \begin{tabular}{lccc}
		\hline
		\hline
    method & \multicolumn{1}{l}{time (sec)} & \multicolumn{1}{l}{cPSNR}  & \multicolumn{1}{l}{shave width}\\
		\hline
    HQLI \cite{HQLI04} & 0.002 $\uparrow$ & 34.34 $\downarrow$  & {4}\\
		HA \cite{HA96}    & 0.024 $\uparrow$ & 33.49 $\downarrow$  &{4}\\
		\rowcolor{DarkGray}
    OS-AP \cite{OSAP10} & 0.04 $\approx$ & 33.26 $\downarrow$  & {10}\\
		\rowcolor{DarkGray}
    LSLCD-NE \cite{LSLCD13} & 0.05 $\approx$  & 34.08 $\downarrow$ & {11}\\
		\rowcolor{DarkGray}
		WECD \cite{WECD06} & 0.13 $\downarrow$ & 32.19 $\downarrow$  & {4}\\
		\rowcolor{DarkGray}
    HPHD \cite{HPHD07} & 0.14 $\downarrow$ & 34.75 $\downarrow$  & {10}\\
		\rowcolor{DarkGray}
		PCSD \cite{PCSD04} & 0.14 $\downarrow$ & 34.93 $\downarrow$ & {3} \\
		\rowcolor{DarkGray}
    AP \cite{AP02}   & 0.29 $\downarrow$ & 33.27 $\downarrow$  & {10}\\
		\rowcolor{DarkGray}
		RA \cite{RA08}  & 0.30 $\downarrow$ & 34.29 $\downarrow$  & {4}\\
		\rowcolor{DarkGray}
    DFAPD \cite{DFAPD07} & 0.45 $\downarrow$ & 34.28 $\downarrow$ & {4}\\
		\rowcolor{DarkGray}
    DFW \cite{DFW12}  & 0.52 $\downarrow$ & 34.58 $\downarrow$ & {6}\\
		\rowcolor{DarkGray}
		WALC \cite{WALC07} & 0.65 $\downarrow$ & 33.85 $\downarrow$ & {4} \\	
		RI  \cite{RI13}  & 0.75 $\downarrow$ & 36.50 $\uparrow$  & {10}\\
		\rowcolor{DarkGray}
		AHD \cite{AHD05}  & 0.97 $\downarrow$ & 33.52 $\downarrow$ & {4} \\
    ICC \cite{ICC14}  & 1.03 $\downarrow$ & 36.79 $\uparrow$ & {10}\\	
		MLRI \cite{MLRI16} & 1.65 $\downarrow$ & 36.91 $\uparrow$ & {10}\\
		CS  \cite{CS12}    & 1.68 $\downarrow$ & 35.59 $\uparrow$ & {4} \\
		\rowcolor{DarkGray}
    DLMMSE \cite{DLMMSE05} & 1.88 $\downarrow$ & 34.40 $\downarrow$  & {20}\\
		\rowcolor{DarkGray}
		SSD \cite{SSD09}   & 4.32 $\downarrow$ & 35.38  $\approx$ & {4}\\
		\rowcolor{DarkGray}
		MSG \cite{MSG13}  & 7.29 $\downarrow$ & 34.72 $\downarrow$ & {10}\\
    DDR \cite{DDR16}  & 7.32 $\downarrow$ & 37.17 $\uparrow$   & {4}\\
		MDWI \cite{MDWI15} & 17.09 $\downarrow$ & 36.16 $\uparrow$ & {10}\\
    ARI \cite{ARI15}  & 25.23 $\downarrow$ & 37.49 $\uparrow$ & {10}\\
		\rowcolor{DarkGray}
		TLS \cite{TLS06}  & 151.08 $\downarrow$ & 30.67 $\downarrow$ & {4}\\
		\rowcolor{DarkGray}
    FlexISP \cite{FlexISP14} & 158.06 $\downarrow$ & 35.45 $\approx$ & {4}\\
    LDINAT \cite{LDINAT11}  & 264.10 $\downarrow$ & 36.18 $\uparrow$  & {20}\\
		\rowcolor{DarkGray}
		SEM \cite{SEM16}  & 568.69 $\downarrow$ & 34.19 $\downarrow$ & {7}\\
		\rowcolor{DarkGray}
    ADMM \cite{ADMM17} & 587.97 $\downarrow$ & 32.25 $\downarrow$ & {4}\\
		LED (ours) & 0.06  & 35.23 & {4}\\
		\hline
    \end{tabular}%
		\vspace{1ex}
		\caption{The mean cPSNR values and median running time (in seconds) of the competing methods and the proposed LED on McM. Down-arrow symbol ``$\downarrow$'' (or up-arrow symbol ``$\uparrow$'') indicates that the corresponding performance is inferior (or superior) to the proposed method; Symbol ``$\approx$'' means similar performance to the proposed method. Methods that are evidently outperformed by our method are highlighted for clarity. }
  \label{tab:LEDvsOthers}%
\end{table*}%

\subsection{Visual Performance on Low Resolution Images}
\begin{figure*}[th!]
	\centering
	\subfigure[Original]{
		\includegraphics[width=0.28\textwidth]{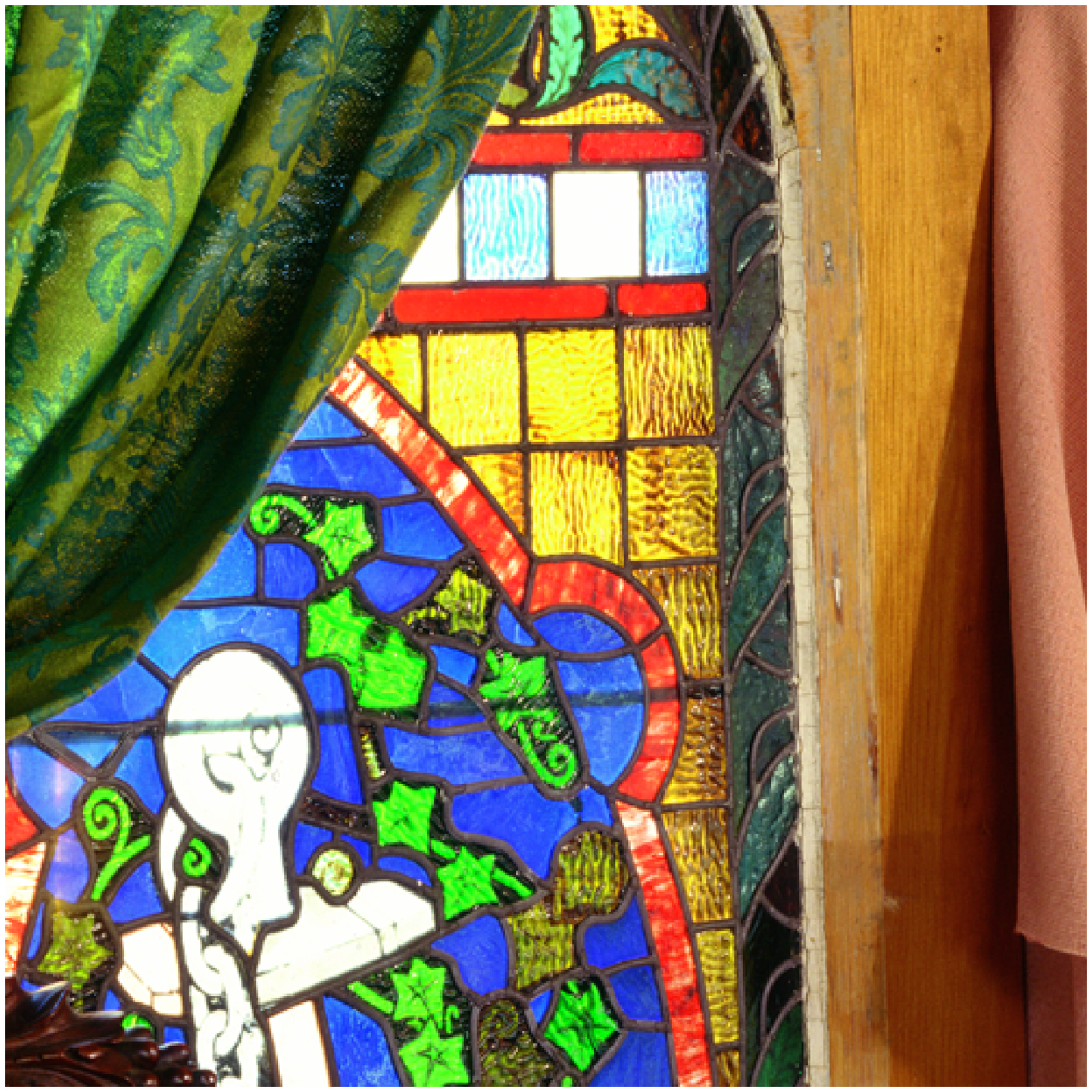}
	}~
	\subfigure[Zoomed]{
		\includegraphics[width=0.28\textwidth]{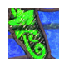}
	}~
	\subfigure[ICC]{
		\includegraphics[width=0.28\textwidth]{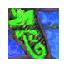}
  }\\
	\subfigure[MLRI]{
		\includegraphics[width=0.28\textwidth]{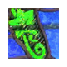}
	}~
	\subfigure[DDR]{
		\includegraphics[width=0.28\textwidth]{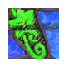}
	}~
	\subfigure[LED (proposed)]{
		\includegraphics[width=0.28\textwidth]{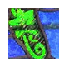}
	}
	\caption{An example of the demosaicking results by ICC, MLRI, DDR and the proposed LED on a local region of McM image 1. Visually, LED performs better than ICC and MLRI.}
	\label{fig:Visual1}
\end{figure*}

\begin{figure*}[t!]
	\centering
	\subfigure[Original]{
		\includegraphics[width=0.28\textwidth]{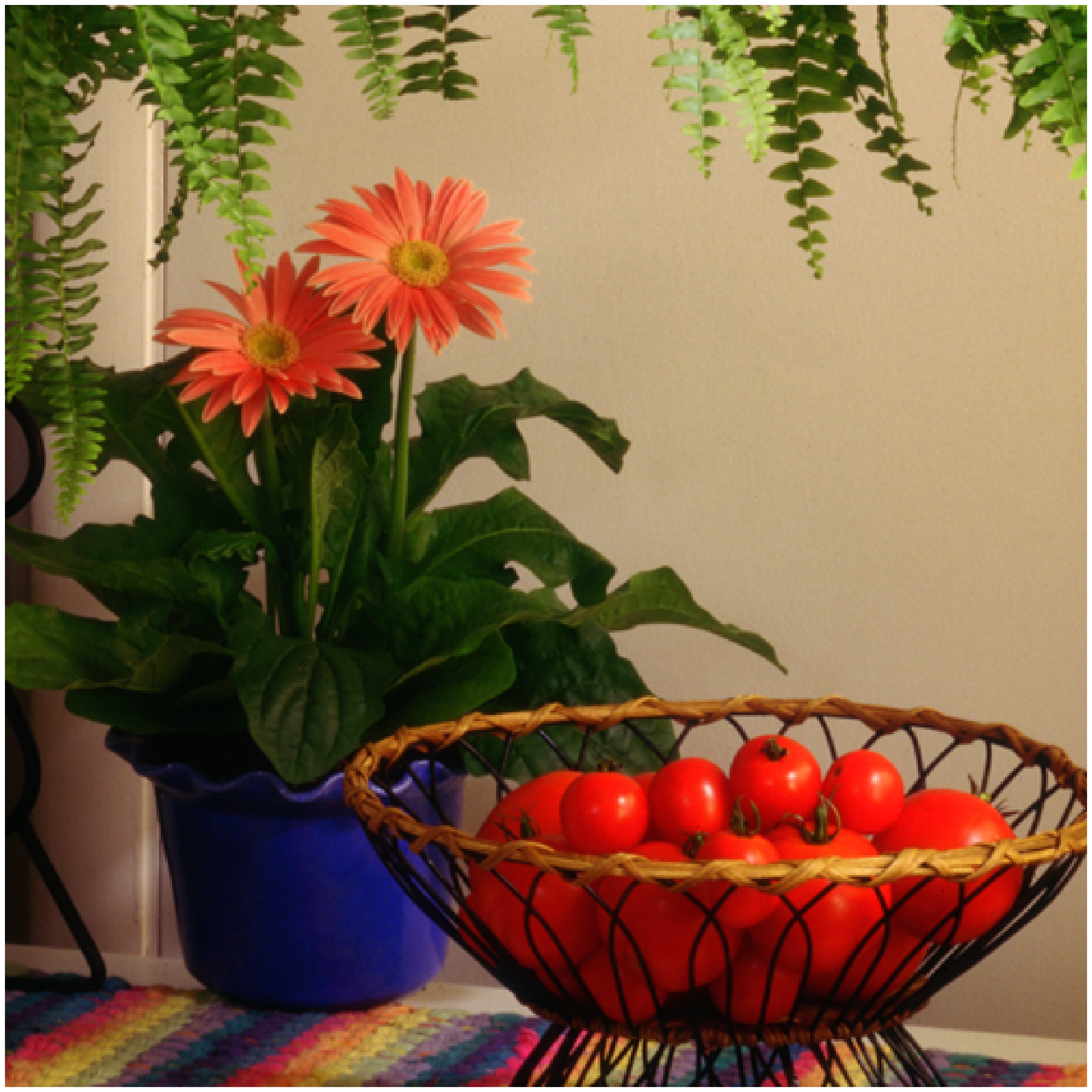}
	}~
	\subfigure[Zoomed]{
		\includegraphics[width=0.28\textwidth]{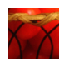}	
	}~
	\subfigure[ICC]{
		\includegraphics[width=0.28\textwidth]{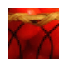}
  }\\
	\subfigure[MLRI]{
		\includegraphics[width=0.28\textwidth]{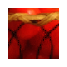}
	}~
	\subfigure[DDR]{
		\includegraphics[width=0.28\textwidth]{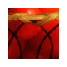}
	}~
	\subfigure[LED]{
		\includegraphics[width=0.28\textwidth]{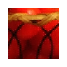}
	}
	\caption{An example of the demosaicking results by ICC, MLRI, DDR and the proposed LED, on a local region of McM image 9. Visually, LED performs better than MLRI.}
	\label{fig:Visual9}
\end{figure*}

\begin{figure*}[th!]
	\centering
	\subfigure[Original]{
		\includegraphics[width=0.29\textwidth]{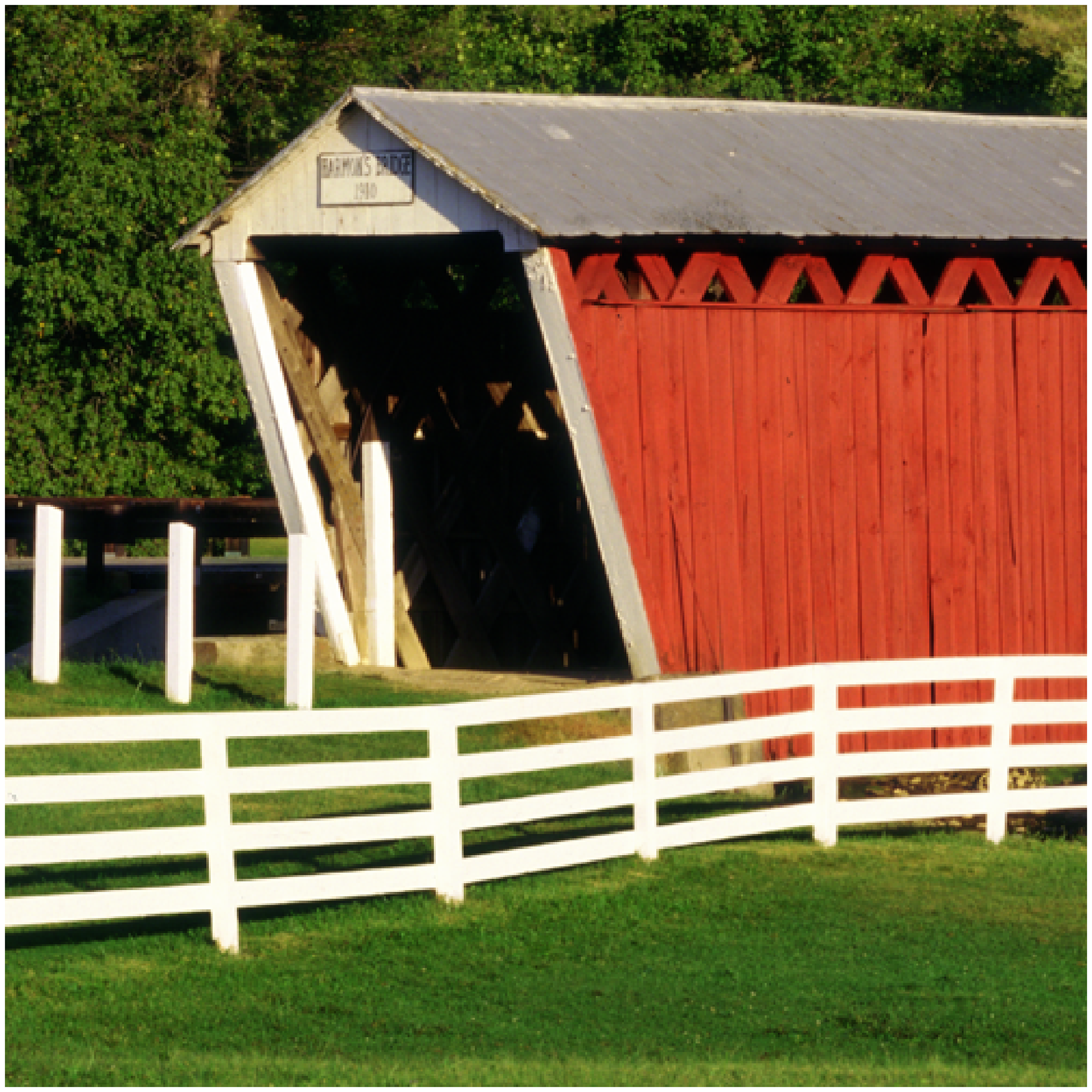}
	}~
	\subfigure[Zoomed]{
		\includegraphics[width=0.29\textwidth]{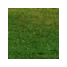}
	}~
	\subfigure[ICC]{
		\includegraphics[width=0.29\textwidth]{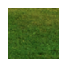}
  }\\
	\subfigure[MLRI]{
		\includegraphics[width=0.29\textwidth]{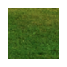}
	}~
	\subfigure[DDR]{
		\includegraphics[width=0.29\textwidth]{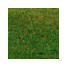}
	}~
	\subfigure[LED (proposed)]{
		\includegraphics[width=0.29\textwidth]{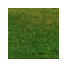}
	}
	\caption{An example of the demosaicking results by ICC, MLRI, DDR and the proposed LED on a local region of McM image 18. Visually, LED performs better than DDR.}
	\label{fig:Visual18}
\end{figure*}

Fig.\ref{fig:Visual1}-Fig.\ref{fig:Visual18} show examples that LED works visually favorably to state-of-the-art methods. Fig.\ref{fig:Visual1} shows a local region taken from the 1st image of McM. Demosaicking by ICC and MLRI in this region suffers noticeable ``false color'' artifacts, whereas DDR and LED recoveries look more natural. Fig.\ref{fig:Visual9} shows another example taken from the 9th McM image. MLRI incompletely recovers the black lines in the example region, whereas LED and ICC both slightly blur the black lines with the red background, but their recovery is visually more acceptable. In the example shown by Fig.\ref{fig:Visual18}, DDR produces obvious ``smearing'' artifacts. In contrast, demosaicking results by ICC, MLRI and LED are all visually close to the original image. 

\subsection{Evaluation on Modern Resolution Images}
The resolution of the MR Kodak dataset is similar to today's popular daily-use cameras, which entail fast demosaicking speed. Table \ref{tab:LEDvsOthersMRKodak} compares the proposed LED with faster algorithms HA and HQLI, as well as more sophisticated algorithms RI, ICC, MLRI, CS and DDR, which have higher cPSNR accuracy than LED on low resolution dataset McM. In this experiment, we exclude methods that are more than $200$ times slower than LED, since they would have different application scenarios. On test images of modern resolution, LED and DDR have the highest average SSIM value. It outperforms CS by cPSNR, SSIM and running time. The cPSNR of LED is still notably (more than 1db) higher than HA and HQLI, while it is comparable to the top-performing state-of-the-art methods that are tens of time slower. It takes LED only $2.86$ seconds, but takes RI, ICC, MLRI tens of seconds and DDR hundreds of seconds. 

\begin{table*}[tp]
  \centering
    \begin{tabular}{ccccccccc}
		\hline
		\hline
    {metrics} & {HQLI} & {HA} & {RI} & {ICC} & {CS} & {MLRI} & {DDR} & {LED}\\
		\hline
		time(sec)   & \textbf{0.034} &  0.820  & 20.96  & 30.87 & 33.71  & 50.21 & 192.90& 2.86\\
		cPSNR       & 41.23 &  39.90  & 42.50  & 42.55 & 41.60  & 42.74 & \textbf{42.79} & 42.28\\
		SSIM        & 0.975 &  0.967  & 0.974  & 0.978 & 0.971  & 0.975 & \textbf{0.980} & \textbf{0.980}\\
		\hline
    \end{tabular}%
		\vspace{1ex}
		\caption{The mean cPSNR, SSIM and median running time of HQLI, HA, RI, ICC, CS, MLRI, DDR and LED demosaicking methods on the MR Kodak dataset. Bold numbers indicate the best performance under each metric.}
  \label{tab:LEDvsOthersMRKodak}%
\end{table*}%

\section{Conclusion}
\label{sec:conclusion}

We have proposed a very low cost edge sensing strategy, termed as LED, for color image demosaicking. It guides the green channel interpolation and color difference plane interpolation by logistic functional of the difference between directional variation. Among 29 demosaicking methods tested by code running, our method is one of the fastest. Without using any refinement or post-processing technique, LED achieves the accuracy higher than many recently proposed methods on low resolution images, and comparable to top performers on images of currently popular resolution. Our extensive experiments suggest that, \emph{accurate non-local edge detection for demosaicking is generally difficult and time consuming. Instead, leveraging the originally captured values of the nearest neighbours is much more efficient.} 

Our algorithm is highly parallelable, and hence its GPU or FPGA implementation can easily restore very high resolution images in real time. This is desirable in the digital camera industry, as the camera resolution is increasing rapidly. Furthermore, in demosaicking applications where speed is allowed to trade for accuracy, the proposed method provides a quick and high quality initialization, which is generally needed in sophisticated iterative demosaicking algorithms.


%
\bibliographystyle{IEEEtran}

 Generated by IEEEtran.bst, version: 1.12 (2007/01/11)

\end{document}